\documentclass[sigconf]{acmart}% anonymous, review]{acmart}
\usepackage[utf8]{inputenc}
\usepackage[T1]{fontenc}
\usepackage{algorithmic}
\usepackage[ruled,vlined]{algorithm2e}
\usepackage{soul} % for strickthrough

\usepackage{caption}
\usepackage{amsmath}

\usepackage[ruled,vlined]{algorithm2e}
\usepackage{soul} % for strikethrough
\usepackage{array}
\usepackage{tabularx}

\usepackage[compact]{titlesec}
    \titlespacing{\section}{0pt}{2ex}{1ex}
    \titlespacing{\subsection}{0pt}{1ex}{0ex}
    \titlespacing{\subsubsection}{0pt}{0.5ex}{0ex}

\usepackage[position=bottom]{subfig}%the position=bottom ensures the captures are at the bottom of the figures
\AtBeginDocument{%
  \providecommand\BibTeX{{%
    \normalfont B\kern-0.5em{\scshape i\kern-0.25em b}\kern-0.8em\TeX}}}

\copyrightyear{2022}
\acmYear{2022}
\setcopyright{rightsretained}
\acmConference[GECCO '22 Companion]{Genetic and Evolutionary Computation Conference Companion}{July 9--13, 2022}{Boston, MA, USA}
\acmBooktitle{Genetic and Evolutionary Computation Conference Companion (GECCO '22 Companion), July 9--13, 2022, Boston, MA, USA}
\acmDOI{10.1145/3520304.3528944}
\acmISBN{978-1-4503-9268-6/22/07}

\begin{document}
\begin{small}
%\title{Geodesics and Non-linearity: Why the Archive Is Important In Novelty Search}
%\title{Geodesics, Biased Mutations And Novelty Search}
\title{Geodesics, Non-linearities and the Archive of Novelty Search}

\author{Achkan Salehi, Alexandre Coninx, Stephane Doncieux}
\affiliation{%
  \institution{Sorbonne Université, CNRS, ISIR, F-75005}
  \city{Paris} 
  \country{France} 
}
\email{{achkan.salehi, alexandre.coninx, stephane.doncieux}@sorbonne-universite.fr}
%\email{alexandre.coninx@sorbonne-universite.fr}

%\email{stephane.doncieux@sorbonne-universite.fr}
%

\renewcommand{\shortauthors}{Salehi et al.}

\begin{abstract}
  The Novelty Search (NS) algorithm was proposed more than a decade ago. However, the mechanisms behind its empirical success are still not well formalized/understood. This short note focuses on the effects of the archive on exploration. Experimental evidence from a few application domains suggests that archive-based NS performs in general better than when Novelty is solely computed with respect to the population. An argument that is often encountered in the literature is that the archive prevents exploration from backtracking or cycling, \textit{i.e.} from revisiting previously encountered areas in the behavior space. We argue that this is not a complete or accurate explanation as backtracking \textemdash beside often being desirable \textemdash can actually be enabled by the archive. Through low-dimensional/analytical examples, we show that a key effect of the archive is that it counterbalances the exploration biases that result, among other factors, from the use of inadequate behavior metrics and the non-linearities of the behavior mapping. Our observations seem to hint that attributing a more active role to the archive in sampling can be beneficial.
\end{abstract}




\maketitle

\section{Introduction}

In recent years, Novelty Search (NS) \cite{lehman2011abandoning, doncieux2019novelty} has increasingly been studied and applied in contexts ranging from robotic manipulation \cite{kim2021exploration} and Reinforcement Learning \cite{jackson2019novelty} to swarm robotics \cite{gomes2013evolution} and games \cite{liapis2015constrained}.  

Traditionally, two different variations of NS have been used: the archive-free one, also called behavioral diversity \cite{mouret2012encouraging}, in which the Novelty score is computed with respect to the population, and the archive-based one, in which the aforementioned score is computed with respect to the archive and the population. Some (limited) evidence \cite{gomes2015devising, cully2017quality, mouret2015illuminating} suggests that the use of an archive is in general beneficial for exploration itself. However, very few explanations have been proposed for theses observations, and they often do not go beyond speculation. An argument that is often encountered in the literature \cite{mouret2015illuminating, lehman2011abandoning, velez2014novelty, salehi2021br} is that the archive prevents backtracking (also called "cycling") in the behavior space: that is, its role is to prevent the exploration from visiting areas that have been previously visited. This is inaccurate or at least incomplete, as while, in fact, avoiding recently visited areas can promote dynamic exploration in the short term, in the longer term, backtracking is often \textit{desirable}: even in the ideal and simplified case where the mapping from genotype space to phenotypes is one-to-one, accessing some areas of behavior space might depend on rare events, whose realization might require to return frequently enough to a particular area of the behavior space. In practice, as illustrated by the low-dimensional analytical problems that we discuss, archives that are large enough to succeed on a given task in general encourage backtracking over longer time horizons.

\begin{figure}[ht!]
  %trim={left,lower,right,up}
  \includegraphics[width=34mm,clip]{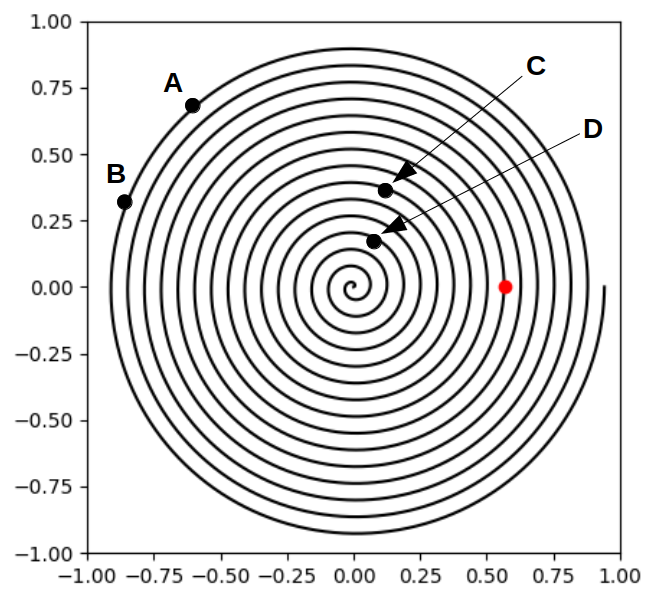}
  \includegraphics[width=34mm,trim={0.1cm 0.5cm 2.5cm 0},clip]{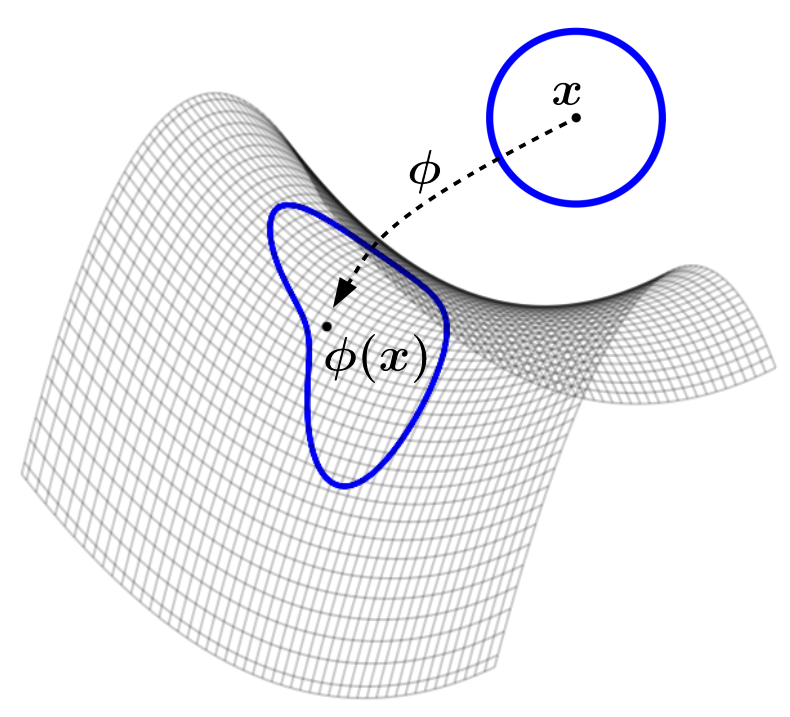}
  \caption{\small{(Left) An Archimedean spiral with parameter $a=0.01$ and $t \in [0, 30\pi]$ embedded in $\mathbb{R}^2$. Using the Euclidean distance and $k=1$, points $A$ and $B$ will have higher Novelty scores than $C$ and $D$. In contrast, using the geodesic distance, $C$ and $D$ would have much higher Novelty than $A$ and $B$. In this figure, the red dot indicates the starting point of the agents. (Right) In practice, even when the reachable behavior space is a smooth manifold, Isotropic Gaussian mutations of a genotype $x$ result in behavior exploration that is locally biased. In this figure, isotropic Gaussian mutations within one standard deviation are mapped to the interior of the bean curve $r=0.7\cos^3(t)^3+\sin^3(t)$ which is then mapped to the surface $2x^2-4y^2$. In more practical cases, \textit{e.g.} with neural network controllers, the resulting distribution will in general be much more complex.}}
  \label{problems_intro}
\end{figure}
\begin{figure*}[h!t!]
  \centering
  %trim={left,lower,right,up}
  %\captionsetup[subfigure]{justification=centering,font=scriptsize}
  \captionsetup[subfigure]{justification=centering}
  %\subfloat[]{
    \subfloat[Euclidean distance, $\phi_b$ parametrization.][Euclidean distance,\\ $\phi_b$ (angle) parametrization.]{
      \includegraphics[width=38mm,trim={1cm 0 1cm 1.4cm},clip]{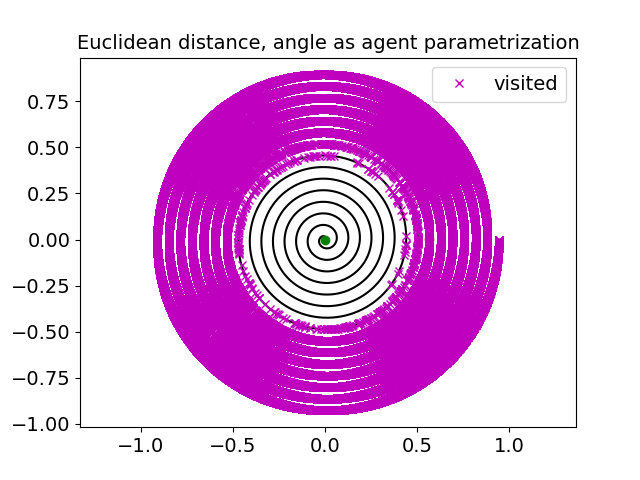}}
    \subfloat[Euclidean distance, $\phi_u$ parametrization.][Euclidean distance,\\ $\phi_u$ (arc-length) parametrization.]{
      \includegraphics[width=38mm,trim={1cm 0 1cm 1.4cm},clip]{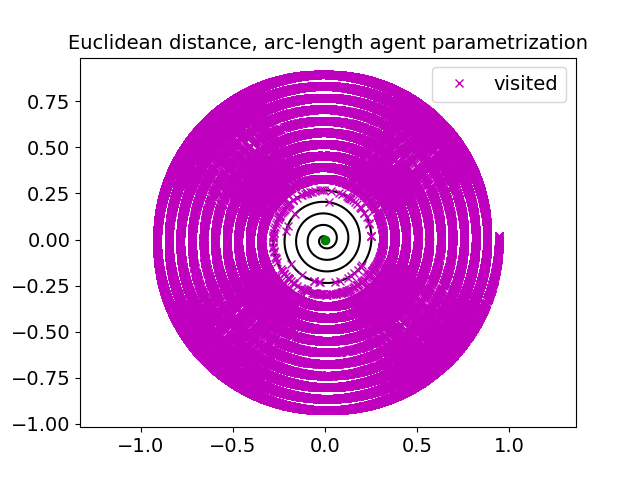}}
    \subfloat[Geodesic distance, $\phi_b$ (angle) parametrization.][Geodesic distance,\\ $\phi_b$ (angle) parametrization.]{
      \includegraphics[width=38mm,trim={1cm 0 1cm 1.4cm},clip]{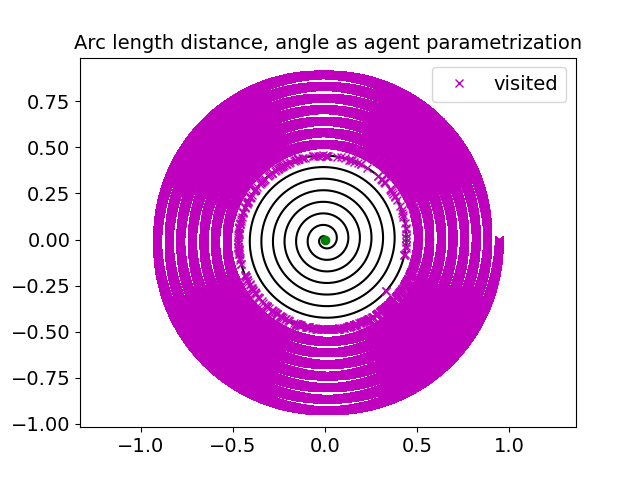}}
    \subfloat[Geodesic distance, $\phi_u$ (arc length) parametrization.][Geodesic distance,\\ $\phi_u$ (arc length) parametrization.]{
      \includegraphics[width=38mm,trim={1cm 0 1cm 1.4cm},clip]{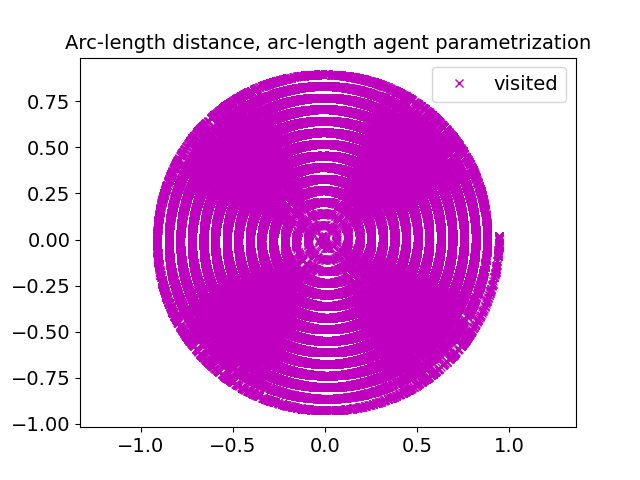}}\\
    \caption{\small{The cumulative results of $20$ Novelty search experiments in the four problem settings. Note that in (a) and (c), a Gaussian distribution centered at $g_b \in \mathcal{G}_b$ is mapped to a distribution with mean $\phi_u(g_b)$ but that is skewed towards the exterior of the spiral. In (a) and (b), the use of the Euclidean metric results in lower Novelty scores towards the interior of the spiral. In these $20$ experiments, Novelty search was performed without an archive, and full behavior space coverage proved to be possible only in the case where the geodesic distance and non-biased mutations were used (d).}}
   \label{spiral_metrics}
\end{figure*}

In this paper, we study the advantages of using an archive for exploration by first inspecting how exploration can be biased by non-linearity and the use of the often inadequate Euclidean metric, which does not account for curvature or discontinuities. We build on low-dimensional, analytical examples, based on which we hypothesize that the usefulness of the archive for exploration stems, in part, from its ability to compensate those biases. While that bias mitigation can naturally occur in bounded behavior spaces with unbounded archives, we observe that this might not necessarily be the case for practical archives with manageable sizes without some improvements to the manner in which archives are defined and used in NS. In particular, those observations, coupled with the fact that individuals that are far in genotype space can in many situations be mapped to (nearly) identical behaviors, indicate that the archive should play a more active role in the sampling process and not be decoupled from the population. \\

\noindent \textbf{Notations and reference algorithms.} Given a genotype space $\mathcal{G}$, we assume that a hand-engineered or learned behavior space $\mathfrak{B}$ has been defined, and denote $\phi:\mathcal{G} \rightarrow \mathfrak{B}$ the mapping between the two spaces. Considering an evolutionary optimization process running for $G_{max}$ generations, the current population at generation $g$ will be noted $\mathcal{P}^g$, and we will write $\mathcal{A}$ the archive that is used with NS. Both unstructured and structured archives have been used in our experiments. In the unstructured case, archive growth was based on the selection of random elements from the population at each generation, and removals were also random. The structured archives that we used were simple grid-like partitionings as in the original MAP-elites algorithm \cite{mouret2015illuminating}, and as we focus on pure exploratory search (\textit{i.e.} no fitness/reward), addition of individuals to each cell was based on its occupancy information: individuals falling in empty cells were immediately added. Otherwise, the individual already occupying the cell was replaced with a small probability $\epsilon$.
The bound imposed on the size of bounded archives will be noted $\mathcal{A}_{max}$.

\section{On biased exploration}
\label{sec_spiral}
%
%It is not just about preventing the algorithm from revisiting where it has gone before. Sometimes, it is even desirable to revisit previously encountered individuals: those might be the 
%
In this section, we place ourselves in the idealized case where the phenotype space is a Riemannian manifold embedded in $\mathbb{R}^n$. We define \textit{locally biased} exploration at a point $b \in \mathfrak{B}$ as an exploration process that favors particular directions in behavior space, \textit{i.e.} a sampling process that is not isotropic. We note that undesired biases might arise from these sources:

\begin{figure*}[ht!]
  \centering
  %trim={left,lower,right,up}
  \captionsetup[subfigure]{justification=centering}
  \subfloat[Unlimited archive][Unlimited archive]{
    \includegraphics[width=36.4mm,trim={1cm 0 1cm 1.4cm},clip]{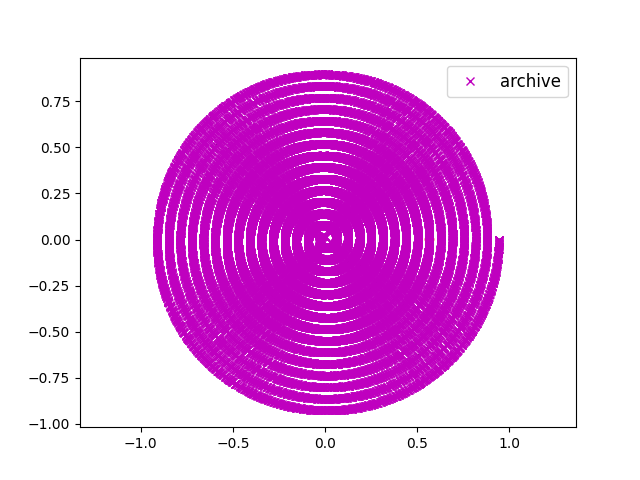}}
    %\subfloat[Unstructured archive, maximum size of 50][Unstructured archive,\\ maximum size of 50]{
    %\includegraphics[width=42mm,trim={1cm 0 1cm 1.4cm},clip]{ims/finite_archive_bad_metrics_50.png}}
  \subfloat[Evolution of the median of selected mutations $\mathcal{H}$, unlimited archive][Evolution of the median of selected mutations $\mathcal{H}$,\\ unlimited archive]
  {\raisebox{2ex}{
    \includegraphics[width=36mm,trim={0.2cm 0 1cm 1.0cm},clip]{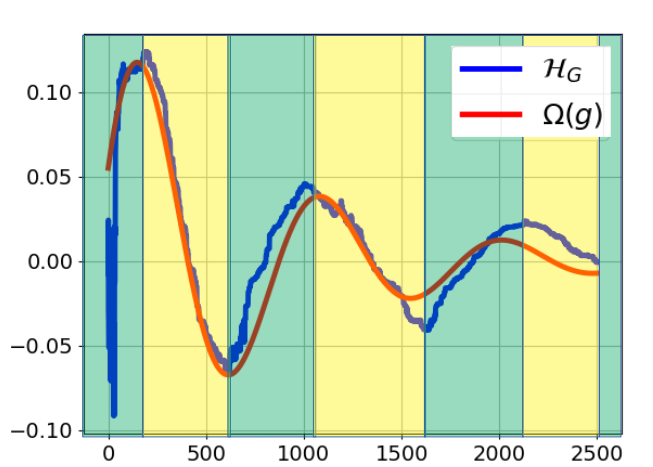}}}
    \subfloat[Unstructured archive, maximum size of $100$][Unstructured archive,\\ maximum size of $100$]{
    \includegraphics[width=38mm,trim={0.6cm 0 1.0cm 1.4cm},clip]{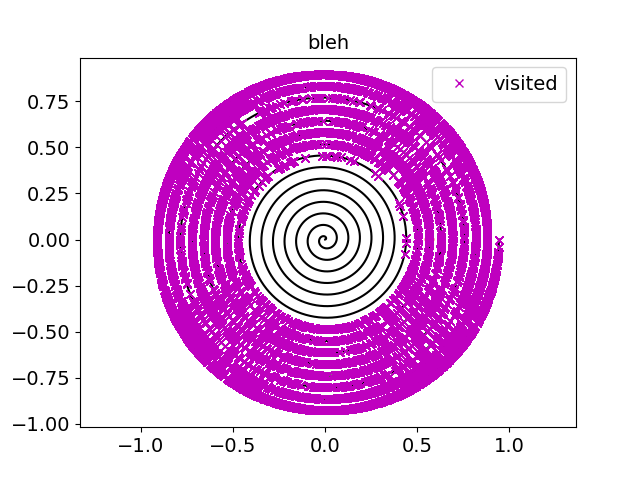}}
  \subfloat[Evolution of the median of selected mutations $\mathcal{H}$, unstructured archive of maximum size $100$][Evolution of the median of selected mutations $\mathcal{H}$, unstructured archive of maximum size $100$]{\raisebox{2ex}{
    \includegraphics[width=38mm,trim={0.2cm 0 0cm 0cm},clip]{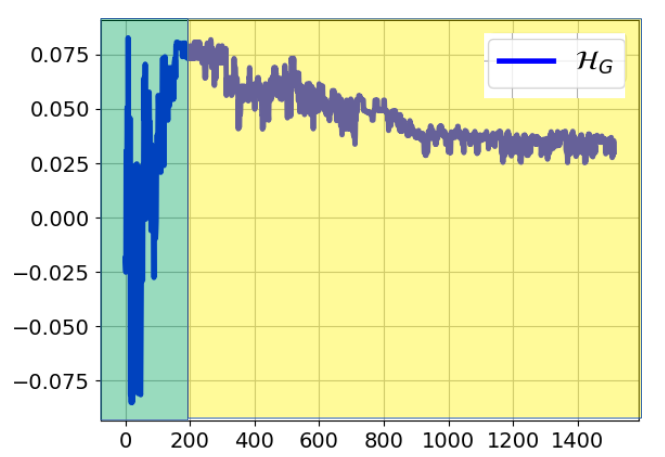}}}

  \subfloat[Unstructured archive, maximum size of 50][Unstructured archive,\\ maximum size of 50]{
    \includegraphics[width=38mm,trim={1cm 0 1cm 1.4cm},clip]{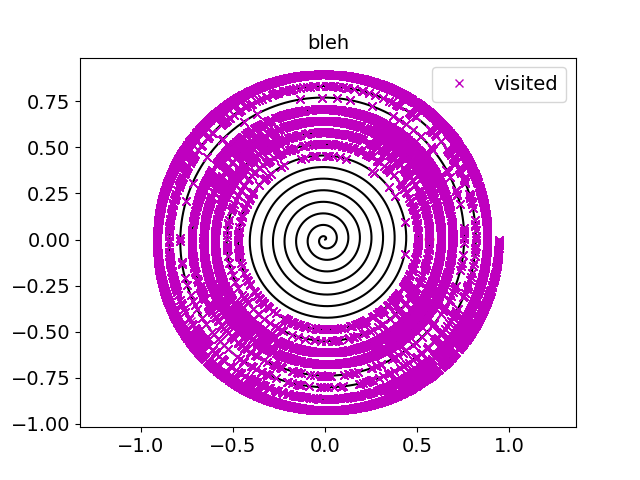}}
    %\subfloat[Unstructured archive, maximum size of 50][Unstructured archive,\\ maximum size of 50]{
    %\includegraphics[width=38mm,trim={1cm 0 1cm 1.4cm},clip]{ims/finite_archive_bad_metrics_50.png}}
    \subfloat[Unstructured archive, maximum size of 200][Unstructured archive,\\ maximum size of 200]{
    \includegraphics[width=38mm,trim={1cm 0 1cm 1.4cm},clip]{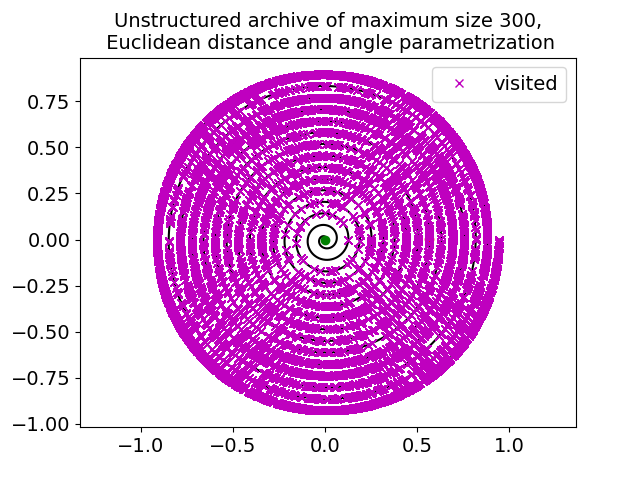}}%%%%%%%%%%%%%%this is 200
    \subfloat[Unstructured archive, maximum size of 3000][Unstructured archive,\\ maximum size of 3000]{
    \includegraphics[width=38mm,trim={1cm 0 1cm 1.4cm},clip]{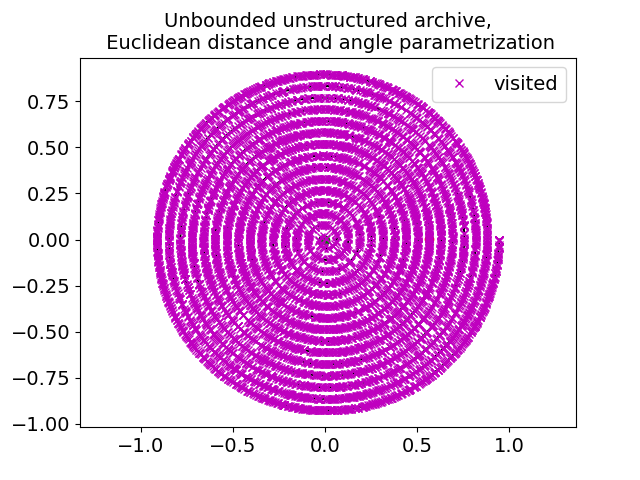}}
    \subfloat[Structured archive,no resampling in the archive][Structured archive,\\ no resampling in the archive]{
    \includegraphics[width=38mm,trim={1cm 0 1cm 1.4cm},clip]{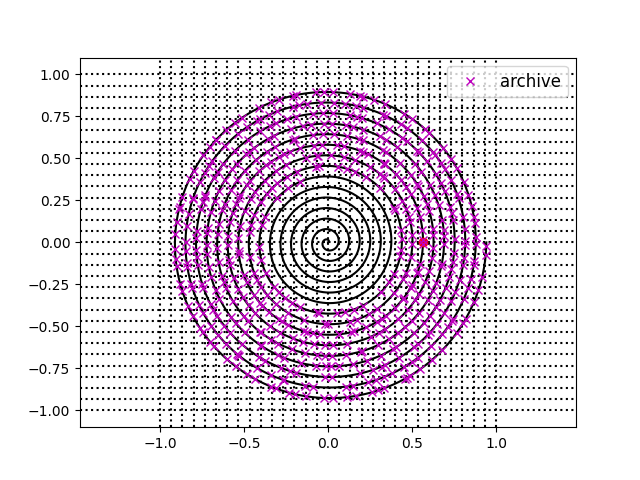}}
    
  \subfloat[Unstructured archive, maximum size of 200 and resampling in the archive][Unstructured archive,\\ maximum size of 200 and\\ random resampling in the archive]{
    \includegraphics[width=38mm,trim={1cm 0 1cm 1.4cm},clip]{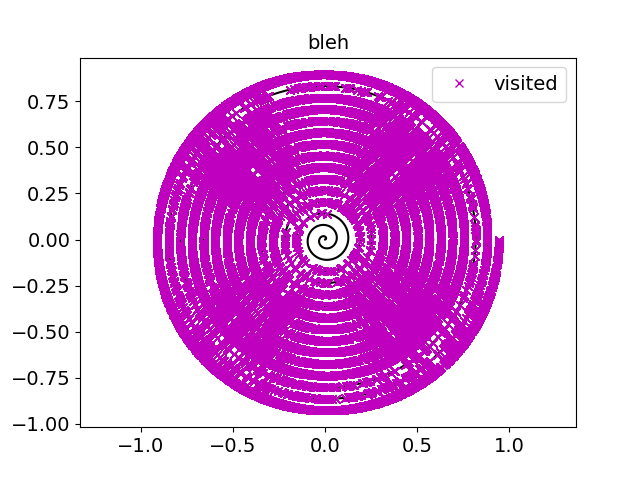}}
  \subfloat[Large population without an archive, same compuational budget as in (i)][Large population without an\\ archive, same compuational budget\\ as in (i)]{
    \includegraphics[width=38mm,trim={1cm 0 1cm 0},clip]{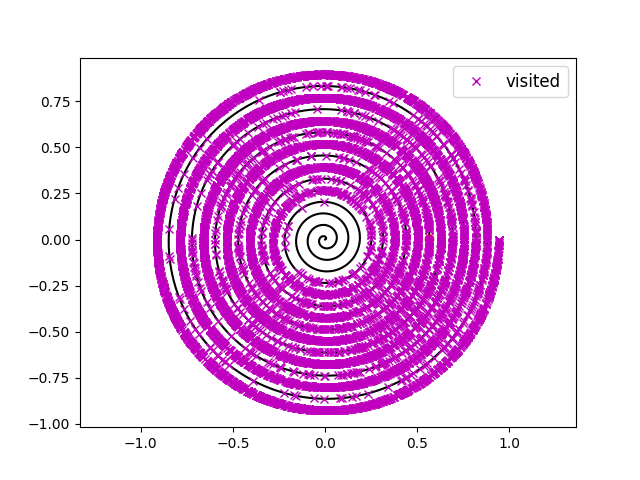}}
     \subfloat[Structured archive,random resampling in the archive][Structured archive,\\ random resampling in the archive]{
    \includegraphics[width=38mm,trim={1cm 0 1cm 0},clip]{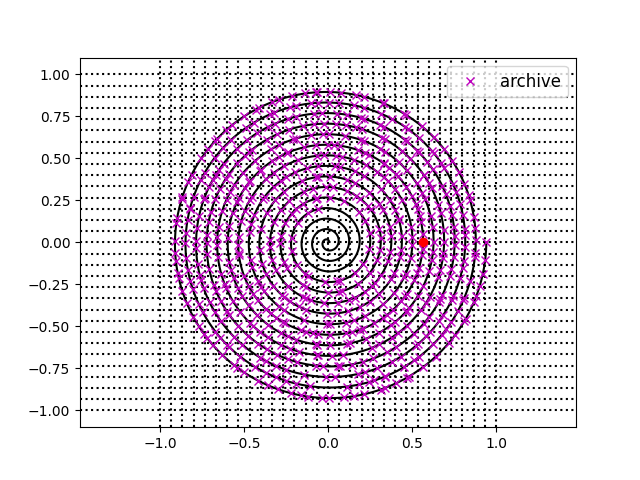}}
  \subfloat[Structured archive, guided resampling in the archive (using $\eta_g$)][Structured archive,\\ guided resampling in the archive (using $\eta_g$)]{
    \includegraphics[width=38mm,trim={1cm 0 1cm 0},clip]{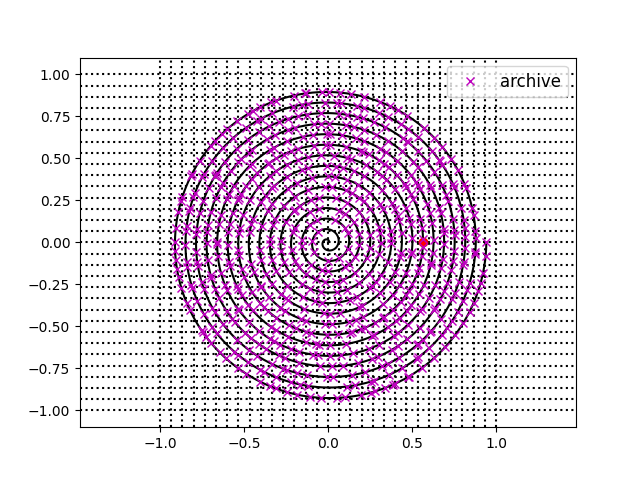}}\\
  \caption{\small{Cumulated results of experiments (with $G_{max}=1000$) obtained with different archive configurations when the Euclidean metric and the $\phi_b$ parametrization were used.}}
   \label{spiral_metrics_archive}
\end{figure*}

\begin{itemize}
  \item The Novelty objective is often computed based on the Euclidean metric\footnote{In very few instances of applying the Novelty objective, such as in the context of robotic manipulation \cite{salehi2021few}, distances are measured in the tangent plane of $SE(N)$. However, the reachable behavior space is limited by the scene and the limits of the actuators, resulting in discontinuities and irregularities that no longer correspond to those metrics.}, which could not only be different but entirely contradictory with Novelty based on geodesic distances, as illustrated in figure \ref{problems_intro} (left).
  \item The non-linearity of the genotype to phenotype mapping $\phi$ results in non-isotropic exploration of the behavior space (figure \ref{problems_intro} (right)). Indeed, even if we assume isotropic gaussian mutations with sufficiently small variance $\sigma$ (a condition that will not hold in practice), the resulting random variable $\phi(x)$ will have covariance $\sigma \frac{\partial{\phi}}{\partial{x}} \frac{\partial{\phi}}{\partial{x}}^T$, which is not isotropic.
\end{itemize}

We hypothesize that balanced exploration of the behavior space requires the mitigation of those biases, and illustrate that point by studying the following toy setting: let us assume a reachable phenotype space $\mathfrak{B}_{toy}$ given by an archimedean spiral (figure \ref{problems_intro}, left) embedded in $\mathbb{R}^2$ whose equation in cartesian coordinates is given by (for $t \in \mathbb{R}^{+}$)

\begin{equation}
  \gamma(t)=(at\cos(t), at\sin(t))
\end{equation}

and for which the geodesic distance naturally coincides with its arc-length, given by

\begin{equation}
  S(t_1,t_2)=\frac{a}{2}(t\sqrt{t^2+1} + \log(t+\sqrt{t^2+1}))\rvert_{t_1}^{t_2}.
\end{equation}

Let us also define the two distinct bounded genotype spaces 

\begin{equation}
  \begin{cases}
    \mathcal{G}_b\triangleq [0,\alpha\pi]\\
    \mathcal{G}_u\triangleq \{S(0,l) \rvert l \in [0, \alpha\pi]\}
  \end{cases}
\end{equation}

for some positive real $\alpha$. Indeed, $\mathcal{G}_b$ corresponds to the angle parametrization of the curve, and $\mathcal{G}_u$ is the set of possible arc-lengths. Let us use an isotropic Gaussian mutation operator, and finally, let us define two mappings from those genotypes to $\mathbb{R}^2$:

\begin{equation}
  \begin{cases}
    \phi_b(g_b)\triangleq \gamma(g_b) & \forall g_b \in \mathcal{G}_b \\
    \phi_u(g_u)\triangleq \gamma(S^{-1}(g_u)) & \forall g_u \in \mathcal{G}_u
  \end{cases}
\end{equation}. 

The motivation behind this choices for $\phi_b$ and $\phi_u$, which both have $\mathfrak{B}_{toy}$ as their image, is that they behave differently under Gaussian mutations: it can easily be verified that an isotropic Gaussian distribution in $\mathcal{G}_b$ will be mapped by $\phi_b$ to a distribution that is skewed towards the exterior of the spiral, while an isotropic Gaussian distribution in $\mathcal{G}_u$ will be mapped by $\phi_u$ to an isotropic Gaussian on $\gamma$. Note that the mapping $S^{-1}$ can not be expressed in closed form. As a result, whenever necessary, we approximate its values by solving the corresponding ordinary differential equation.

These settings define four problems, that result from pairing one of the two parametrizations $\mathcal{G}_b$ or $\mathcal{G}_u$ with one of the two metrics (Euclidean or Geodesic $S(t_1,t_2)$). We expect that using the geodesics distance in conjunction with $\mathcal{G}_u$ should result in balanced exploration, and that all other cases NS will heavily favor the exploration to move towards the exterior of the spiral. We verified this hypothesis by running $20$ experiments corresponding to each of those four problems, setting $\alpha=30$, with population and offspring sizes of $M=N=30$, mutations sampled from $\mathcal{N}(0,0.3)$ and the consideration of $k=10$ nearest neighbors. In all of the experiments, the number of generations was set to $G_{max}=1000$, and all individuals were initialized with a particular value of $s$ (the red dot in figure \ref{problems_intro} (left)). The cumulated results, that are reported in figure \ref{spiral_metrics} confirm our hypothesis: the only outcome in which the cumulative exploration results fully cover the behavior space (figure \ref{spiral_metrics} (d)) corresponds to the case where the geodesic distance is used with $\mathcal{G}_u$. In figures \ref{spiral_metrics}(a) and (c), exploration seems to be trapped around the external borders of the behavior space. Figure \ref{spiral_metrics} (b) seems to indicate that using a parametrization $\mathcal{G}_u$ that does not induce bias is not sufficient to compensate the selection pressure that results from the use of the wrong metric.

This finally brings us to the archive. As we will discuss in the next section, the use of a large unstructured archive can compensate the aforementioned biases in the long term, precisely \textit{because it encourages back-tracking}, without the need for learning the correct (\textit{i.e.} geodesic) metric or designing mutation operators that do not induce bias (both of which are impractical in real-world problems).

\section{Bias compensation using an archive}
\label{section_bias_comp_archive}

The experiments presented throughout the section use the bias-inducing parametrization $\mathcal{G}_b$, and the Euclidean metric. Unless explicitly stated, the same settings for $\alpha, M, N, k, G_{max}$ as in the previous section are used. The same holds for mutation standard deviations. 

\subsection{Large unstructured archives can overcome exploratory biases by enabling backtracking} With bounded behavior spaces, it is intuitively expected that an unbounded archive will eventually be able to overcome any biases in the selection pressures, and that is indeed the case for the spiral (\ref{spiral_metrics_archive} (a)). To explicit the effects of the archive on selection pressures, let us consider the history of of all selected mutations, over $G$ generations. More precisely, denote by $\psi(.)$ the operator that maps an individual in the population $\mathcal{P}^{g}$ to its parent in $\mathcal{P}^{g-1}$, and define 

\begin{equation}
h_g\triangleq \{S(0,p)-S(0,\psi(p))\}_{p \in \mathcal{P}^g}.
\end{equation}

In other terms, each element of $h_g$ is the behavioral change, expressed as a change in arc length, that results from a mutation to some individual $p$ at generation $g$. Now, consider the history of the medians of all such behavioral mutations over $G$ generations $\mathcal{H}_G \triangleq \{\text{median}(h_i)\}_{i=0}^{G-1}$, which is displayed in figure \ref{spiral_metrics_archive} (b) (blue curve). We see the emergence of a pattern reminiscent of a \textit{damped harmonic oscillatory} motion, as illustrated by the fitted curve that is displayed in red in figure \ref{spiral_metrics_archive}(b), and which converges to $0$ in term of amplitude as $g$ increases. Interestingly, each of its half-periods correspond to either an \textit{expansion} or \textit{retraction} stage in exploration:  each expansion stage (shaded in green), corresponds to a set of consecutive generations during which the population consistently moves towards the exterior of the spiral. Likewise, each retraction stage, shaded in yellow, corresponds to a set of consecutive generations during which the population is pushed towards the center of the spiral. It can be seen that the amplitude (\textit{i.e.} the median of the selected mutations) approaches $0$ as $g$ grows, indicating that asymptotically, the use of an unbounded archive does indeed lead to unbiased, balanced exploration.

This pattern seems to hint that in this settings \textemdash and using the population-based sampling of traditional Novelty search \textemdash, a bounded archive will fail to cover the entire space unless it is sufficiently large to push the exploration first through an expansion phase before \textit{backtracking} towards the center. This is confirmed by the experiments with archive sizes of $\mathcal{A}_{max}=50,100,200$ (figure \ref{spiral_metrics_archive}(c,d,e,f)). In particular, it is clear from \ref{spiral_metrics_archive}(d) that those archive sizes fail because they are not large enough to encourage a complete backtracking phase. In our experiments, it appeared that in order to ensure success over a single run consistently it was necessary that $\mathcal{A}_{\max}>1000$. An example for $\mathcal{A}_{\max}=3000$ is given in figure \ref{spiral_metrics_archive}(g). 

Structured archives fail to cover the behavior space in $G_{max}$ generations without significant modification to the sampling process (figures \ref{spiral_metrics_archive}(h, k)). This is expected, since in that case, sampling only from the population eliminates the possibility of backtracking (figure \ref{spiral_metrics_archive}(h)) and allowing sampling from the archive as in MAP-elites (figure \ref{spiral_metrics_archive}(k)) converges too slowly: backtracking requires the selection of a single particular cell, which happens too infrequently. We recognize that this might be a particularity of the low-dimensional problem at hand. Indeed, while only a single path between two points $b_0, b_1 \in \mathfrak{B}_{toy}$ exists on the spiral, in higher dimensions, unless we consider degenerate cases, an infinity of such paths would be available. However, we note the notorious difficulty of choosing the adequate partitioning and resolution of a structured archive which is also present in higher dimensions.

\subsection{Discussion}
We saw that non-linearities and discrepancies between the distance metric and the local structure of the behavior space were a source of bias in exploration, and showed that it could be compensated by using an unstructured archive that was large enough to enable backtracking past the initialization point. As the time complexity of NS with an unstructured archive is $O(|\mathcal{A}|\log|\mathcal{A}|)$ \cite{mouret2015illuminating}, in practice, bounds should be imposed on its size. Consequently, a logical step would be to modify the sampling strategy \textemdash which by default samples the population \textemdash to facilitate backtracking. However, as illustrated by figure \ref{spiral_metrics_archive}(i,k), simply allowing resampling from the archive (as in classical MAP-elites) leads to unreasonably slow convergence as progress relies on the sampling of rare events. As the reader has probably noticed, the situation in which we allow resampling into an unstructured archive is to some extent similar, conceptually, to Novelty search without an archive but with a larger population (keeping the same computational budget). As is illustrated in figure \ref{spiral_metrics_archive}(j), the latter performs worse than the former (results aggregated over $5$ runs). This is not surprising from an intuitive point of view, as archive-less NS does not impose the additional local selection pressures that result from archive densifications.

We hypothesize that a more reliable sampling approach is to augment NS with a selection pressure that is \textit{absolute}, in the sense that it is relative to the phenotype space $\mathfrak{B}$ rather than relative to an ever-moving unstructured archive. For example, considering NS with a structured archive, let us assign a score $\eta_g$, that we dub \textit{temporary discovery score}, to each individual $p$ from the parent population $\mathcal{P}^g$:

\begin{equation}
  \eta_g(p)=\tau \eta_{g-1}(p) + (1-\tau)\frac{\sum_{q \in \xi_{p}}\kappa(q)}{\sum_{p\in\mathcal{P}^g}\sum_{q \in \xi_{p}}\kappa(q)},
\end{equation}

where $\xi_{p}$ denotes the set of offsprings from $p$ at generation $g$, and $\kappa(q)$ is simply set to whether the cell in which $\phi(q)$ falls is empty or not. The parameter $\tau \in [0,1]$ sets the rate at which this score is updated with new information. As is shown in figure \ref{spiral_metrics_archive}(l), combining the population-based sampling of NS with resampling in the archive guided by the $\eta_g$ score consistently results in highly improved coverage of the behavior space in all executions.

Note that the suggested score can be thought of as a more flexible version of the Curiosity score proposed in \cite{cully2017quality} in the context of Quality-Diversity. Indeed, the latter is discrete and penalizes the lack of new discoveries by the same amount by which it rewards new behaviors. In contrast, the $\tau$ parameter of the $\eta_g$ score allows for balancing between positive rewards and penalties. 

We close this section with the remark that a possible way in which $\kappa$ could be defined in the context of NS with unstructured archives is to use a loss based on Random Network Distillation \cite{burda2018exploration}. However, that is conditional to whether or not one is able to mitigate the catastrophic inference issues \cite{pfulb2018catastrophic} that are well-known in the context of continuous learning with encoders.
\section*{ACKNOWLEDGMENT}
This work was supported by the European Union's H2020-EU.1.2.2 Research and Innovation Program through FET Project VeriDream under Grant Agreement Number 951992.

\bibliographystyle{ACM-Reference-Format}
\bibliography{sample-bibliography} 


\begin{thebibliography}{15}

%%% the Web address.

\ifx \showCODEN    \undefined \def \showCODEN     #1{\unskip}     \fi
\ifx \showDOI      \undefined \def \showDOI       #1{#1}\fi
\ifx \showISBNx    \undefined \def \showISBNx     #1{\unskip}     \fi
\ifx \showISBNxiii \undefined \def \showISBNxiii  #1{\unskip}     \fi
\ifx \showISSN     \undefined \def \showISSN      #1{\unskip}     \fi
\ifx \showLCCN     \undefined \def \showLCCN      #1{\unskip}     \fi
\ifx \shownote     \undefined \def \shownote      #1{#1}          \fi
\ifx \showarticletitle \undefined \def \showarticletitle #1{#1}   \fi
\ifx \showURL      \undefined \def \showURL       {\relax}        \fi
% invisible to TeX
\providecommand\bibfield[2]{#2}
\providecommand\bibinfo[2]{#2}
\providecommand\natexlab[1]{#1}
\providecommand\showeprint[2][]{arXiv:#2}

\bibitem[\protect\citeauthoryear{Burda, Edwards, Storkey, and Klimov}{Burda
  et~al\mbox{.}}{2018}]%
        {burda2018exploration}
\bibfield{author}{\bibinfo{person}{Yuri Burda}, \bibinfo{person}{Harrison
  Edwards}, \bibinfo{person}{Amos Storkey}, {and} \bibinfo{person}{Oleg
  Klimov}.} \bibinfo{year}{2018}\natexlab{}.
\newblock \showarticletitle{Exploration by random network distillation}.
\newblock \bibinfo{journal}{\emph{arXiv preprint arXiv:1810.12894}}
  (\bibinfo{year}{2018}).
\newblock


\bibitem[\protect\citeauthoryear{Cully and Demiris}{Cully and Demiris}{2017}]%
        {cully2017quality}
\bibfield{author}{\bibinfo{person}{Antoine Cully} {and}
  \bibinfo{person}{Yiannis Demiris}.} \bibinfo{year}{2017}\natexlab{}.
\newblock \showarticletitle{Quality and diversity optimization: A unifying
  modular framework}.
\newblock \bibinfo{journal}{\emph{IEEE Transactions on Evolutionary
  Computation}} \bibinfo{volume}{22}, \bibinfo{number}{2}
  (\bibinfo{year}{2017}), \bibinfo{pages}{245--259}.
\newblock


\bibitem[\protect\citeauthoryear{Doncieux, Laflaqui{\`e}re, and
  Coninx}{Doncieux et~al\mbox{.}}{2019}]%
        {doncieux2019novelty}
\bibfield{author}{\bibinfo{person}{Stephane Doncieux}, \bibinfo{person}{Alban
  Laflaqui{\`e}re}, {and} \bibinfo{person}{Alexandre Coninx}.}
  \bibinfo{year}{2019}\natexlab{}.
\newblock \showarticletitle{Novelty search: a theoretical perspective}. In
  \bibinfo{booktitle}{\emph{Proceedings of the Genetic and Evolutionary
  Computation Conference}}. \bibinfo{pages}{99--106}.
\newblock


\bibitem[\protect\citeauthoryear{Gomes, Mariano, and Christensen}{Gomes
  et~al\mbox{.}}{2015}]%
        {gomes2015devising}
\bibfield{author}{\bibinfo{person}{Jorge Gomes}, \bibinfo{person}{Pedro
  Mariano}, {and} \bibinfo{person}{Anders~Lyhne Christensen}.}
  \bibinfo{year}{2015}\natexlab{}.
\newblock \showarticletitle{Devising effective novelty search algorithms: A
  comprehensive empirical study}. In \bibinfo{booktitle}{\emph{Proceedings of
  the 2015 Annual Conference on Genetic and Evolutionary Computation}}.
  \bibinfo{pages}{943--950}.
\newblock


\bibitem[\protect\citeauthoryear{Gomes, Urbano, and Christensen}{Gomes
  et~al\mbox{.}}{2013}]%
        {gomes2013evolution}
\bibfield{author}{\bibinfo{person}{Jorge Gomes}, \bibinfo{person}{Paulo
  Urbano}, {and} \bibinfo{person}{Anders~Lyhne Christensen}.}
  \bibinfo{year}{2013}\natexlab{}.
\newblock \showarticletitle{Evolution of swarm robotics systems with novelty
  search}.
\newblock \bibinfo{journal}{\emph{Swarm Intelligence}} \bibinfo{volume}{7},
  \bibinfo{number}{2} (\bibinfo{year}{2013}), \bibinfo{pages}{115--144}.
\newblock


\bibitem[\protect\citeauthoryear{Jackson and Daley}{Jackson and Daley}{2019}]%
        {jackson2019novelty}
\bibfield{author}{\bibinfo{person}{Ethan~C Jackson} {and} \bibinfo{person}{Mark
  Daley}.} \bibinfo{year}{2019}\natexlab{}.
\newblock \showarticletitle{Novelty search for deep reinforcement learning
  policy network weights by action sequence edit metric distance}. In
  \bibinfo{booktitle}{\emph{Proceedings of the Genetic and Evolutionary
  Computation Conference Companion}}. \bibinfo{pages}{173--174}.
\newblock


\bibitem[\protect\citeauthoryear{Kim, Coninx, and Doncieux}{Kim
  et~al\mbox{.}}{2021}]%
        {kim2021exploration}
\bibfield{author}{\bibinfo{person}{Seungsu Kim}, \bibinfo{person}{Alexandre
  Coninx}, {and} \bibinfo{person}{St{\'e}phane Doncieux}.}
  \bibinfo{year}{2021}\natexlab{}.
\newblock \showarticletitle{From exploration to control: learning object
  manipulation skills through novelty search and local adaptation}.
\newblock \bibinfo{journal}{\emph{Robotics and Autonomous Systems}}
  \bibinfo{volume}{136} (\bibinfo{year}{2021}), \bibinfo{pages}{103710}.
\newblock


\bibitem[\protect\citeauthoryear{Lehman and Stanley}{Lehman and
  Stanley}{2011}]%
        {lehman2011abandoning}
\bibfield{author}{\bibinfo{person}{Joel Lehman} {and}
  \bibinfo{person}{Kenneth~O Stanley}.} \bibinfo{year}{2011}\natexlab{}.
\newblock \showarticletitle{Abandoning objectives: Evolution through the search
  for novelty alone}.
\newblock \bibinfo{journal}{\emph{Evolutionary computation}}
  \bibinfo{volume}{19}, \bibinfo{number}{2} (\bibinfo{year}{2011}),
  \bibinfo{pages}{189--223}.
\newblock


\bibitem[\protect\citeauthoryear{Liapis, Yannakakis, and Togelius}{Liapis
  et~al\mbox{.}}{2015}]%
        {liapis2015constrained}
\bibfield{author}{\bibinfo{person}{Antonios Liapis},
  \bibinfo{person}{Georgios~N Yannakakis}, {and} \bibinfo{person}{Julian
  Togelius}.} \bibinfo{year}{2015}\natexlab{}.
\newblock \showarticletitle{Constrained novelty search: A study on game content
  generation}.
\newblock \bibinfo{journal}{\emph{Evolutionary computation}}
  \bibinfo{volume}{23}, \bibinfo{number}{1} (\bibinfo{year}{2015}),
  \bibinfo{pages}{101--129}.
\newblock


\bibitem[\protect\citeauthoryear{Mouret and Clune}{Mouret and Clune}{2015}]%
        {mouret2015illuminating}
\bibfield{author}{\bibinfo{person}{Jean-Baptiste Mouret} {and}
  \bibinfo{person}{Jeff Clune}.} \bibinfo{year}{2015}\natexlab{}.
\newblock \showarticletitle{Illuminating search spaces by mapping elites}.
\newblock \bibinfo{journal}{\emph{arXiv preprint arXiv:1504.04909}}
  (\bibinfo{year}{2015}).
\newblock


\bibitem[\protect\citeauthoryear{Mouret and Doncieux}{Mouret and
  Doncieux}{2012}]%
        {mouret2012encouraging}
\bibfield{author}{\bibinfo{person}{J-B Mouret} {and}
  \bibinfo{person}{St{\'e}phane Doncieux}.} \bibinfo{year}{2012}\natexlab{}.
\newblock \showarticletitle{Encouraging behavioral diversity in evolutionary
  robotics: An empirical study}.
\newblock \bibinfo{journal}{\emph{Evolutionary computation}}
  \bibinfo{volume}{20}, \bibinfo{number}{1} (\bibinfo{year}{2012}),
  \bibinfo{pages}{91--133}.
\newblock


\bibitem[\protect\citeauthoryear{Pf{\"u}lb, Gepperth, Abdullah, and
  Kilian}{Pf{\"u}lb et~al\mbox{.}}{2018}]%
        {pfulb2018catastrophic}
\bibfield{author}{\bibinfo{person}{Benedikt Pf{\"u}lb},
  \bibinfo{person}{Alexander Gepperth}, \bibinfo{person}{S Abdullah}, {and}
  \bibinfo{person}{A Kilian}.} \bibinfo{year}{2018}\natexlab{}.
\newblock \showarticletitle{Catastrophic forgetting: still a problem for DNNs}.
  In \bibinfo{booktitle}{\emph{International conference on artificial neural
  networks}}. Springer, \bibinfo{pages}{487--497}.
\newblock


\bibitem[\protect\citeauthoryear{Salehi, Coninx, and Doncieux}{Salehi
  et~al\mbox{.}}{2021a}]%
        {salehi2021br}
\bibfield{author}{\bibinfo{person}{Achkan Salehi}, \bibinfo{person}{Alexandre
  Coninx}, {and} \bibinfo{person}{Stephane Doncieux}.}
  \bibinfo{year}{2021}\natexlab{a}.
\newblock \showarticletitle{BR-NS: an archive-less approach to novelty search}.
\newblock \bibinfo{journal}{\emph{arXiv preprint arXiv:2104.03936}}
  (\bibinfo{year}{2021}).
\newblock


\bibitem[\protect\citeauthoryear{Salehi, Coninx, and Doncieux}{Salehi
  et~al\mbox{.}}{2021b}]%
        {salehi2021few}
\bibfield{author}{\bibinfo{person}{Achkan Salehi}, \bibinfo{person}{Alexandre
  Coninx}, {and} \bibinfo{person}{Stephane Doncieux}.}
  \bibinfo{year}{2021}\natexlab{b}.
\newblock \showarticletitle{Few-shot Quality-Diversity Optimisation}.
\newblock \bibinfo{journal}{\emph{arXiv preprint arXiv:2109.06826}}
  (\bibinfo{year}{2021}).
\newblock


\bibitem[\protect\citeauthoryear{Velez and Clune}{Velez and Clune}{2014}]%
        {velez2014novelty}
\bibfield{author}{\bibinfo{person}{Roby Velez} {and} \bibinfo{person}{Jeff
  Clune}.} \bibinfo{year}{2014}\natexlab{}.
\newblock \showarticletitle{Novelty search creates robots with general skills
  for exploration}. In \bibinfo{booktitle}{\emph{Proceedings of the 2014 Annual
  Conference on Genetic and Evolutionary Computation}}.
  \bibinfo{pages}{737--744}.
\newblock


\end{thebibliography}

\end{small}
\end{document}